\documentclass[conference]{IEEEtran}
\IEEEoverridecommandlockouts
\usepackage{cite}
\usepackage[utf8]{inputenc}
\usepackage{graphicx}
\usepackage{amsmath}
\usepackage[version=4]{mhchem}
\usepackage{siunitx}
\usepackage{longtable,tabularx}
\usepackage{subcaption}
\usepackage{epstopdf}
\usepackage{tabularx}
\usepackage{booktabs} 
\usepackage{multirow}
\usepackage{arydshln} 
\usepackage{url}
\usepackage{float}
\usepackage{acronym}
\usepackage{fancyhdr}
\usepackage{graphics} 
\usepackage{epsfig} 
\usepackage{amsmath} 
\usepackage{amssymb}  
\usepackage{tabularx}
\usepackage{booktabs} 
\usepackage{multirow}
\usepackage{siunitx}
\usepackage{lipsum} 
\usepackage{bm}
\usepackage{todonotes}
\usepackage{subcaption}
\usepackage{url}
\usepackage{graphicx}
\usepackage{subcaption}
\usepackage{mwe}
\usepackage{textcomp} 
\usepackage{tabularx,multirow,array}
\usepackage{threeparttable}
\usepackage{arydshln}

\DeclareCaptionLabelFormat{leftside}{#1#2\hfill}
\newcommand{\edit}[1]{\textcolor{black}{#1}}
\newcommand{\editt}[1]{\textcolor{blue}{#1}}

\fancypagestyle{specialfooter}{%
  \fancyhf{}
  
  \fancyfoot[C]{1}
}
\usepackage{hyperref} 
\usepackage{textcomp} 
\def\BibTeX{{\rm B\kern-.05em{\sc i\kern-.025em b}\kern-.08em
    T\kern-.1667em\lower.7ex\hbox{E}\kern-.125emX}}

\renewcommand{\footnoterule}{%
  \kern -3pt
  \hrule width \columnwidth height 1pt
  \kern 2pt
}

\begin{document}

\title{Optimizing Design and Control of Running Robots Abstracted as Torque Driven Spring Loaded Inverted Pendulum (TD-SLIP) }

\makeatletter
\newcommand{\linebreakand}{%
  \end{@IEEEauthorhalign}
  \hfill\mbox{}\par
  \mbox{}\hfill\begin{@IEEEauthorhalign}
}
\makeatother
\author{\IEEEauthorblockN{Reed Truax$^1$$^\dagger$ \thanks{$^1$ Ph.D. Student, Department of Mechanical and Aerospace Engineering}}
\IEEEauthorblockA{\textit{University at Buffalo}\\
Buffalo, New York, 14260 \\
reedtrua@buffalo.edu}
\and
\IEEEauthorblockN{Feng Liu$^1$$^\dagger$ \thanks{$^\dagger$ Joint First Author}}
\IEEEauthorblockA{\textit{University at Buffalo}\\
Buffalo, New York, 14260 \\
fliu23@buffalo.edu}
\and
\IEEEauthorblockN{Souma Chowdhury$^2$ \thanks{$^2$ Associate Professor, Mechanical and Aerospace Engineering; Adjunct Associate Professor, Computer Science and Engineering}}
\IEEEauthorblockA{\textit{University at Buffalo}\\
Buffalo, New York, 14260 \\
soumacho@buffalo.edu}
\and
\IEEEauthorblockN{Ryan St. Pierre$^3$$^*$ \thanks{$^3$ Assistant Professor, Mechanical and Aerospace Engineering; Adjunct Assistant Professor, Computer Science and Engineering} \thanks{$^*$ Corresponding Author, ryans@buffalo.edu} \thanks{This work is accepted at the IDETC/CIE2024 Forum.} \thanks{Copyright \copyright 2024 ASME. Personal use of this material is permitted. Permission from ASME must be obtained for all other uses, in any current or future media, including reprinting/republishing this material for advertising or promotional purposes, creating new collective works, for resale or redistribution to servers or lists, or reuse of any copyrighted component of this work in other works}}
\IEEEauthorblockA{\textit{University at Buffalo}\\ Buffalo, New York, 14260 \\ ryans@buffalo.edu}
}

\maketitle
\thispagestyle{specialfooter}

\begin{abstract}
Legged locomotion shows promise for running in complex, unstructured environments. Designing such legged robots requires considering heterogeneous, multi-domain constraints and variables, from mechanical hardware and geometry choices to controller profiles. However, very few formal or systematic (as opposed to ad hoc) design formulations and frameworks exist to identify feasible and robust running platforms, especially at the small (sub 500 g) scale. This critical gap in running legged robot design is addressed here by abstracting the motion of legged robots through a torque-driven spring-loaded inverted pendulum (TD-SLIP) model, and deriving constraints that result in stable cyclic forward locomotion in the presence of system noise. Synthetic noise is added to the initial state in candidate design evaluation to simulate accumulated errors in an open-loop control. The design space was defined in terms of morphological parameters, such as the leg properties and system mass, actuator selection, and an open loop voltage profile. These attributes were optimized with a well-known particle swarm optimization solver that can handle mixed-discrete variables.  Two separate case studies minimized the difference in touchdown angle from stride to stride and the actuation energy, respectively.  Both cases resulted in legged robot designs with relatively repeatable and stable dynamics, while presenting distinct geometry and controller profile choices.
\end{abstract}



\section{Introduction}

Legged robots use their legs and actuators to push against the ground, accelerating and decelerating their bodies to run over terrain. While utilizing wheels, instead of legs, can be easier for design and control, legged robots have the advantage of being able to traverse complex terrain and jump over obstacles, compared to a similarly sized wheeled robot. Although legs offer the ability to dynamically shift contact and traverse complex environments, the design and control of legged robots can be challenging~\cite{kim2017design}. \textbf{\textit{These challenges compound as the scale of the system decreases, placing constraints on sensing, computation, and actuation options~\cite{flynn_gnat,stpierre2019toward}.}} 

Addressing these challenges requires careful joint consideration of the morphology (geometry and component choices) and the behavior (controls) of autonomous legged systems. A critical consideration therein is the requirement to find optimized parameters to return stable and cyclic gaits~\cite{silva2012literature}. One approach is to mimic biological runners by creating robots that are dynamically similar. For example, the relative stiffness, a non-dimensionalized stiffness term in legged locomotion models, must be designed to be between 7 and 30~\cite{monopod,holmes2006dynamics, SHEN2015433} for legged runners.  The relative stiffness can be used to inform design decisions by specifying relationships between the system's mass, leg stiffness, and leg length.  In addition to designing the physical robot parameters, a motor must be selected that can satisfy the speed and torque requirements necessary to maintain a stable gait.  Once all mechanical properties are selected, a controller must be designed to ensure stable and cyclic gaits.  This controller must be designed to accommodate the stride frequency of the system, which scales in proportion to the size and spring-mass ratio of the robot~\cite{Full1990,Heglund1988SpeedSF}.

These design and control challenges in robotics often place cyclic constraints on one another through choices in components. For example, requiring larger motors to generate larger torques requires larger batteries to supply that power.  In turn, this requires more torque, and thusly a larger motor. These cyclic constraints require concurrent design frameworks that consider all component and control choices, as well as coupled constraints~\cite{censi2015mathematical,censi2016class}.

Co-design has been successfully used in a range of legged robotic applications across a number of morphologies such as: single bipedal walkers, legged hoppers, and multi-legged robots. The design variables used in these works vary, with some just considering mass distribution, leg properties, or actuator selection while others look at the entire system design to accomplish specific tasks.  In the area of bipedal walkers, \cite{7862383} examines mass distribution across a number of body morphologies in simulation, and \cite{8202228} expands on this by using co-design to select actuators in addition to mass distribution. Fadini et al. used co-design to select actuators and gearing while minimizing the energy of a hopper \cite{9560988}.  Yesilevskiy et al. expanded the design parameters by considering mass distribution, actuator type, and actuation placement while minimizing the cost of transport for a hopping robot~\cite{7487269}.  In \cite{9349280}, parameters and an open loop controller for the monopod hopping robot Skippy were selected while considering motor dynamics which would maximize jump height and travel distance. Diguarti et al. used co-design to select leg mass and linkage lengths along with a controller that would achieve specific gait types in StarlETH \cite{Digumarti2014}. Reference \cite{9981641} had similar design parameters as in \cite{Digumarti2014} but formulated in a general sense for all four-legged robots.

Most of the above mentioned works consider legs with multiple joints.  While these have the benefit of adjustable leg stiffness and allow for the center of mass of the robot to be controlled with six degrees of freedom, they add additional control and design complications that are not realizable at small scales (\SI{10}{\gram}-\SI{500}{\gram}). This observation leads to the guiding question of the work presented in this paper -- i.e., \textbf{\textit{how optimization based co-design can be leveraged to help provide design guidance at smaller scales.}}

However, while optimization can uniquely help in exploring design choices and trade-offs well beyond what is conceivable even by domain experts, it gives rise to a few challenges when designing complex or novel robotic systems in abstract forms. Expressing the desired capabilities of a novel bio-inspired robotic system (such as the running robot considered here) in a quantitative form amenable to optimization is far from trivial. Firstly, this calls for an understanding of what set of constraint functions are needed to describe the feasible behavior of the system at the conceptual abstracted stage. Secondly, it calls for the imposition of bounds on components (e.g., actuators) and geometric choices that make the system practically realizable. Third, it demands the identification of control architectures (inner loop computations) and the optimization method (outer loop search) that enable a computationally efficient co-design process; not to mention, the outer loop search is likely to present a mixed-integer non-linear programming or MINLP problem (where component choices/features are discrete and geometric choices are continuous). 


This paper makes the following specific contributions to address the above stated challenges and present a computationally efficient framework for morphology/control concurrent design of a small-scale (sub 500 g) running robot that has a stable gait, and is energy efficient: \textbf{1)} We develop a simulation of the running legged concept, by combining a torque driven spring loaded inverted pendulum model with a leg stiffness estimation model and an open-loop control system (for the abstracted DC motor) that switches between the flight and stances phases of the running motion. \textbf{2)} We formulate a novel set of constraints to collectively capture the symmetry of the stance phase, repeatability of the periodic forward motion (w/o having to expensively compute many steps per candidate design), adherence to the assumed spring-loaded inverted pendulum abstraction of the system and comparability with biological running systems. \textbf{3)} We adopt a mixed-discrete Particle Swarm Optimization approach \cite{mdpso,tong2016multi}, a well-known MINLP solver, to present and analyze (observably distinct) optimized design trade-offs that respectively minimize average energy consumption and difference in the touchdown angle (expresses motion repeatability).  

\section{Model Formulation \label{sec:model_formulation}} 
The dynamics of legged locomotion can be abstracted to a minimal model of a point mass atop a spring leg. This model, the spring-loaded inverted pendulum (SLIP) model, has been useful for describing the running locomotion dynamics of organisms across a range of mass scales, from gram-scale cockroaches to humans (\SI{70}{\kilo\gram})~\cite{monopod}. \edit{Despite its simplicity, the SLIP model has been used successfully to describe the gaits of biological systems~\cite{10.1242/jeb.01177, Griffin:2000vp,YANG2022116727, 10.1242/jeb.130.1.155} and has been used to design robotics ranging in size and morphology~\cite{Rhex,Altendorfer_2000, c-quad, barragan2018minirhex, EDUBOT, Goldberg_2017,DASH}. }Here, we include both actuation and dissipation in the SLIP model, formulated as a torque-driven damped SLIP (TD-SLIP), similar to~\cite{stpierre2020viscoelastic,energy_regulation}. 

\begin{figure}
    \centering
    \includegraphics[width=1\linewidth]{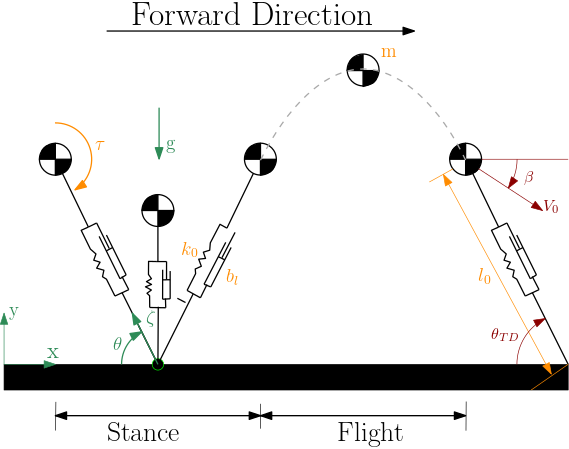}
    \caption{A schematic torque-driven damped spring-loaded inverted pendulum (TD-SLIP) model, with a hip torque and damper parallel to the spring, showing one cycle of stance and flight. Cartesian coordinates $x$, $y$ are used as global coordinates and $\theta$, $\zeta$ are used for locally defining the leg position during stance.}
    \label{fig:FBD}
    \vspace{-15pt}
\end{figure}
\subsection{TD-SLIP Model}
The TD-SLIP model, like the SLIP model, is a cyclic hybrid dynamical model with two phases: a stance phase when the leg is in contact with the ground and a flight phase when the leg is not in contact with the ground. The events of liftoff and touchdown occur between these two phases to mark the transitions in the hybrid dynamics. The model switches from stance to flight in liftoff when the leg reaches its natural length \edit{and has a $y$ acceleration greater than gravity the system} returns to stance in touchdown when the leg makes contact with the ground. 
 
Using the convention defined in Figure \ref{fig:FBD}, the equations of motion are written in polar coordinates for the stance phase as: 
\begin{equation} \label{SLIP_ang} \ddot{\theta}=-\frac{2\dot{\zeta}\theta}{\zeta}-\frac{gcos(\theta)}{\zeta}+\frac{\tau}{m\zeta^2} \end{equation}
\begin{equation} \label{SLIP_z} \ddot{\zeta}=\zeta\dot{\theta}^2-gsin(\theta)-\frac{k_0}{m}(\zeta-l_0)-\frac{b_l}{m}\dot{\zeta} \end{equation}
where $\theta$ and $\zeta$ represent the leg angle and leg length. Table~\ref{tab:sys_param} lists the system parameters associated with the TD-SLIP model. It should be noted that in this formulation, any dissipation from contacting the ground is neglected, and a point contact is assumed. The flight phase is treated as ballistic motion, with the center of mass moving without energy loss, and can be expressed as \(\ddot{x}=0\) and \(\ddot{y}=-g\).  Where $x$ and $y$ represent the horizontal and vertical body positions in Cartesian coordinates. 

In the model, the input torque is provided by a DC motor. While motors are often represented as first-order systems with a mapping between input voltage and speed or output torque, the full second-order representation is used here. The full second-order representation of the model allows for more fidelity in the motor model. For example, this allows us to monitor current draw demands, which would dictate battery discharge rate requirements in hardware. The equations of motion \edit{describing the motors current $i_a$ and rotational speed $\omega$} are expressed as:
\begin{equation} \label{motor_cur} L_a\frac{\delta i_a}{\delta t}=V_a-R_ai_a-k_b\omega \end{equation}
\begin{equation} \label{DC_motor_torq} J\dot{\omega}=\tau-c\omega-\tau_L=k_Ti_a-c\omega-\tau_L \end{equation}

DC motors often require a gearbox to provide the torques and speeds necessary for robot mechanisms. Here, we consider that the motors need a gearbox. \edit{The relationship between the motor shaft speed ($\omega$) and the rotation of the leg ($\dot{\theta}$) is $\omega=R\dot{\theta}$ where $R$ is the gear ratio.}  $J$ must then be modified to account for the inertial effects of the motor shaft and the gearbox, $J=J_m+J_{GB}$. Combining the equations for the TD-SLIP model, with the DC motor equations, requires solving for the \edit{load torque} $(\tau_L)$ in equation~\eqref{DC_motor_torq}, and substituting $\tau_L$ for the \edit{applied torque} $\tau$ in equation~\eqref{SLIP_ang}. This gives the final equation of motion during stance with the geared DC motor:
\begin{equation} \label{theta_fin} \ddot{\theta}(1+\frac{R^2J}{m\zeta^2})=-\frac{2\dot{\zeta}\theta}{\zeta}-\frac{g\cos(\theta)}{\zeta}-\frac{cR^2\dot{\theta}}{m\zeta^2}+\frac{k_ti_aR}{m\zeta^2} \end{equation}
\begin{equation} \label{z_fin} \ddot{\zeta}=\zeta\dot{\theta}^2-g\sin(\theta)-\frac{k_0}{m}(\zeta-l_0)-\frac{b_l}{m}\dot{\zeta} \end{equation}
Equations \eqref{theta_fin} and \eqref{z_fin} along with the equation for current \eqref{motor_cur} are used to describe the dynamics of the system. 
\subsection{Modeling the Leg as a Linear Spring}

\begin{figure}
\centering
\includegraphics[width=0.35\linewidth]{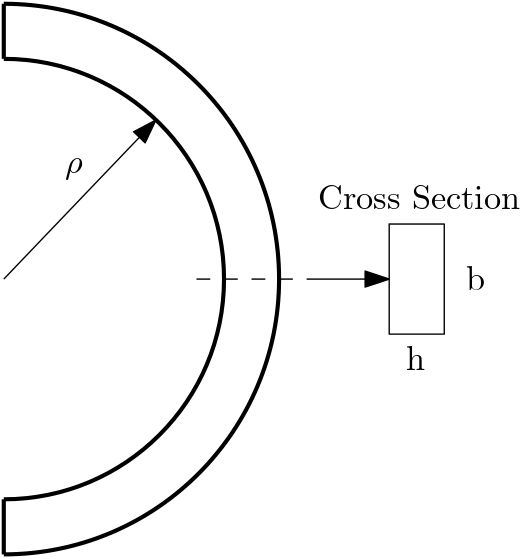}
\caption{Sketch of the leg geometry used for simulation.  The leg is represented as a semicircle with a rectangular cross-section.}
\label{leg_cross_section}
\end{figure}

While the TD-SLIP model abstracts the leg as a linear spring, translating this lumped parameter spring to a mechanical design can be challenging. For this work, we consider the same C-shaped leg with a rectangular cross-section seen in robots across scales, like RHex~\cite{Rhex,saranli2001rhex}, EduBot \cite{EDUBOT}, X-RHex~\cite{galloway2010x}, Mini-RHex~\cite{barragan2018minirhex}, C-Quad \cite{c-quad}, and the robots in~\cite{stpierre2016gait,stpierre2020viscoelastic}.


The stiffness of the C-shaped leg was calculated using Castigliano's theorem to approximate the linear elastic deflection of the leg under a load. Therefore, the stiffness of the leg can be modeled as a linear elastic spring of the form $F=k_0\delta$, and the stiffness of the C-shaped leg is:
\begin{equation} 
\label{castigliano}  
k_0=\frac{bh^3E}{6\rho^3 \pi}
\end{equation}
where the geometric and material properties of the leg, detailed in Table~\ref{tab:opt_param}, will dictate the overall stiffness of the leg. These parameters are used within the optimization framework outlined in Section~\ref{sec:optimization_methods} to guide future hardware implementation.

\newcommand{\STAB}[1]{\begin{tabular}{@{}c@{}}#1\end{tabular}}

\begin{table}

\caption{System Parameters}
\label{tab:sys_param}
\begin{center}
\begin{tabular}{ p{0.35 cm} c| c| l }
\multicolumn{2}{c}{Parameter}   & Units & Explanation\\
\hline
\multicolumn{1}{l}{\multirow{6}{*}{\STAB{\rotatebox[origin=c]{90}{Robot}}}}
& $g$    & \unit{m/s^2} & Gravity \\
\multicolumn{1}{l}{}& $\tau$ & \unit{N-m}   & Input torque \\
\multicolumn{1}{l}{}& $m$    & \unit{kg}    & Mass \\
\multicolumn{1}{l}{}& $k_0$  & \unit{N/m}   & Spring constant, \eqref{castigliano}\\
\multicolumn{1}{l}{}& $l_0$   & \unit{m}     & Initial leg length\\
\multicolumn{1}{l}{}& $b_l$  & \unit{N.s/m} & Linear damping\\
\hline

\multicolumn{1}{l}{\multirow{ 11}{*}{\STAB{\rotatebox[origin=c]{90}{Actuator}}}}
& $V_a$  & \unit{V}      & Applied voltage\\ 
\multicolumn{1}{l}{}& $i_a$  & \unit{A}      & Applied current\\
\multicolumn{1}{l}{}& $R_a$  & \unit{\Omega} & Motor terminal resistance\\
\multicolumn{1}{l}{}& $L_a$  & \unit{H}      & Motor terminal inductance\\
\multicolumn{1}{l}{}& $k_b$  & \unit{N.m/\sqrt{W}}& Back EMF constant\\
\multicolumn{1}{l}{}& $k_t$  & \unit{N.m/A}       & Motor Torque Constant $k_t \equiv k_b$\\
\multicolumn{1}{l}{}& $c$    & \unit{N.m/(rad/s)} & Motor damping\\
\multicolumn{1}{l}{}& $\tau$    & \unit{N.m}      & Torque applied by motor\\
\multicolumn{1}{l}{}& $\tau_L$  & \unit{N.m}      & Load torque\\
\multicolumn{1}{l}{}& $J$    & \unit{kg.m^2}      & Shaft and gearbox inertia\\
\multicolumn{1}{l}{}& $R$    & -      & Gear ratio\\
\hline
\multicolumn{1}{l}{\multirow{ 4 }{*}{\STAB{\rotatebox[origin=c]{90}{Leg}}}}
& $\rho$ & \unit{m}  & Leg radius $l_0=2\rho$ \\
\multicolumn{1}{l}{}& $b$    & \unit{m}  & Leg thickness\\
\multicolumn{1}{l}{}& $h$    & \unit{m}  & Leg width \\
\multicolumn{1}{l}{}& $E$    & \unit{Pa} & Elastic Modulus\\

\end{tabular}
\end{center}
\end{table}
\subsection{Voltage Control}
In hardware, the motors will be actuated with a\edit{n open-loop} time-varying voltage. This control strategy has been successfully used in robots, such as RHex, where the time-varying voltage results in the legs rotating slowly during the stance phase and faster during the flight phase~\cite{Rhex,saranli2001rhex,bclock}. To understand the nominal voltage profile given our model formulation, the TD-SLIP model was simulated over a stance cycle. Figure \ref{fig:com_trajectory} shows how the scaled center of mass trajectory evolves with time across a single stance phase.  Using the rotational speed of the leg ($\dot{\theta}$) and the calculated hip torque at each time step, a voltage profile shown by Figure \ref{fig:com_trajectory} was calculated using equations \eqref{motor_cur}, and \eqref{DC_motor_torq}.  This estimated voltage represents the voltage profile required to match the system's speed and torque requirements at each time.  While the voltage profile is well represented by a third-order system, it does not necessarily represent the most energy-efficient option, which is one of the objectives studied in section \ref{sec:optimization_methods}. 

\begin{figure}

    \centering
    \includegraphics[width=.9\linewidth]{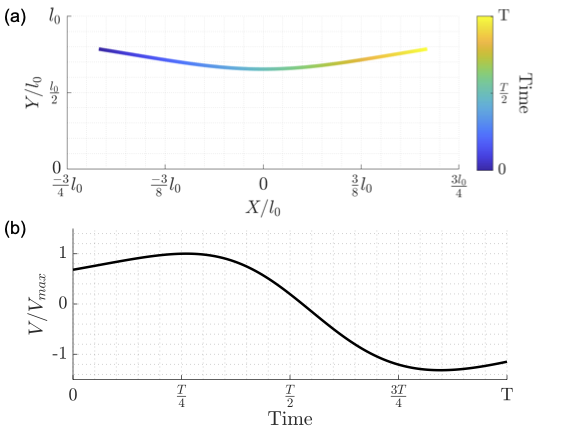}  
  \caption{Panel (a) shows how the trajectory of the center of mass of a single stable stance phase evolves with time.  The $x$ and $y$ coordinates were scaled by $l_0$.  Panel (b) shows the scaled voltage profile required by a motor to match the torque evaluated at the hip and rotational speed during the stance phase shown in panel (a).}  
  \label{fig:com_trajectory}
  \vspace{-15pt}
\end{figure}



To provide more flexibility to the optimization, the voltage profile was expressed as a fifth-order system.  A fifth-order system allows for another voltage directional change during the stance phase, if needed, or can be collapsed to a lower order model.  Therefore, the time-varying voltage will be represented through the constants $a_5$ to $a_0$ and can be expressed as:
\vspace{-5pt}
\begin{equation} 
\label{stanceV} V=a_5t^5+a_4t^4 + \cdots + a_1t+a_0
\vspace{-5pt}
\end{equation}



With a DC motor as the method of actuation, the robot model will continuously rotate the leg, preparing for the next stance during flight. In this formulation, the touchdown angle is able to vary from step to step rather than assuming a constant touchdown angle \edit{as was done in~\cite{non_linear_k,Geyer:2002,Rummel_2010}.  The leg position during flight is controlled through a bang-on bang-off voltage profile, whose objective is to reposition the leg such that the touchdown angle ($\theta_{TD}$) is held constant step to step.  The control voltage is taken to be the maximum voltage rating of the motor used in the system and is applied for a period, $T_{FC}$, the time to reposition the leg to the prescribed touchdown angle.  Section \ref{sec:optimization_methods} outlines motor options in detail, but all options are rated to be run at \SI{3}{V}. Using the maximum voltage, the motor is rated for causes the leg to reposition as quickly as possible.  While this is not necessarily the most energy-efficient option, it guarantees the leg will be in position before touchdown, assuming the motor and gearbox are sized correctly.  With this control scheme, the control method switches at the events of take-off and touchdown.  After detecting an event, the controller will execute the appropriate control profile, which is a function of time.}

\subsection{Simulation Framework}

All equations were simulated using the ode45 function in MATLAB with a variable time step, starting in the stance phase. An event function was used to switch between the differential equations of stance and flight while maintaining continuity. All source code is given in \cite{GitHub_sim}.


\section{Optimization Framework \label{sec:optimization_methods}}



This work considers two co-design case studies using the TD-SLIP framework.  First, co-design is used to design a system that maximizes the symmetry of the first stride through minimizing the change in touchdown angle between the first two cycles.  For a robot operating under open-loop control, any error in the system's dynamics from the control will accumulate from cycle to cycle until the system eventually becomes unstable.  This error can come from the lack of symmetry in the stance cycle and from the repositioning of the leg during flight, preventing repeatable stance cycles.  An error in repositioning will cause the dynamics of a future step to vary from the previous step, causing the dynamics to deviate from their marginally stable gait.  Therefore, it is essential for the robot to minimize the change in the touchdown angle cycle in the presence of uncertainty.  This first case study aims to find a set of design and control parameters that allow the robot to achieve repeatable stance cycles when such accumulation errors exist.

During locomotion the motor is constantly adding energy into the system to propel the body forward and to compensate for energy loss due to damping within the motor and leg.  As batteries can only provide a finite amount of energy, being conservative with power usage is critical to maximizing the endurance of the robot.  Energy usage is distributed across actuation, computation, and sensing. However,  actuation generally consumes the highest amounts of energy and power compared to sensing and computation. This work just considers the contribution of actuation, as was done in ~\cite{DBLP:journals/corr/abs-2103-04660,9560988}. By minimizing energy consumption, the robot is capable of executing a longer mission and possesses the potential for more complex missions. Therefore, a second case study is defined to find the optimized design and control variables that maintain stable and repeatable gaits with minimal energy consumption.  Finding the optimized energy consumption is also done in the presence of noise in the initial touchdown angle.

\subsection{Case Study 1: Optimizing Touchdown Angle Difference}



To represent an accumulated error in the touchdown angle, a Gaussian-distributed noise is added to the initial touchdown angle $(\theta_0)$. Thus, the initial cycle within the simulation can be viewed as a continuation of a preceding, uninterrupted sequence of cycles, representing an ongoing process rather than a beginning of a standalone new cycle. The standard deviation of the Gaussian noise, \(\epsilon\), is defined to be \(1.290^{\circ}\) which will result in a \(\pm 3^{\circ}\) noise with a 98\% confidence interval.  
Therefore, the initial touchdown angle after adding the noise is defined as:
\begin{equation}
\label{eq:epsilon}
    \theta_{\text{TD1}} = \theta_{0} + \mathcal{N}(0, \epsilon^2)
\end{equation}
For a sequence of cycles to be repeatable and stable, the initial touchdown angle among each cycle should be similar, which means after a flight phase, the robot can return to its initial touchdown angle; this ensures that after a flight phase, the robot can return to its initial touchdown angle, allowing it to repeat the previous motion when the same control mechanism is applied. Thus, the objective of the optimization is to minimize the touchdown angle difference between the first stance phase and the second stance phase in the simulation. Here, by definition, the touchdown angle of the second stance phase is the same as the angle between the robot leg and the ground at the end of the first cycle. The formulation of the single objective optimization is defined in equation \eqref{eq:obj_mdpso_angle_diff}.

\begin{equation}
\label{eq:obj_mdpso_angle_diff}
\begin{aligned}
\min_{\mathbf{X}} \quad &  \theta_{\text{Diff}}(\mathbf{X}) = |\theta_{\text{0}} - \theta_{\text{TD2}}(\mathbf{X})|\\
{s. t.}\quad & \mathbf{X}\in[\mathbf{X}_{\text{L}},\mathbf{X}_{\text{U}}]\\
\quad & \text{Touchdown angle constraints: }\\
& \begin{array}{ll}
g_1 = \min (\theta_{\text{TD1}}, \theta_{\text{TD2}}) > 0.45, & g_2 = \theta_{\text{Diff}} < \epsilon, \\
g_3 =  \max (\theta_{\text{TD1}}, \theta_{\text{TD2}}) < 1.48  &  \\
\end{array} \\
\quad & \text{Position constraints: }\\
& \begin{array}{ll}
g_4 = y_{\text{s1\_M}} < 0.85 y_{\text{s1\_S}}, & g_5 = y_{\text{s1\_M}} < 0.85 y_{\text{s1\_E}},\\
g_6 = x_{\text{s1\_E}} > 0, & g_7 = \min (\mathbf{y}_{\text{s1}}) > 0 \\
g_{8} = |\Delta x_{\text{f1}}| - 4 \cdot l_0 < 0 , & g_{9} = \min (\mathbf{y}_{\text{s2}}) > 0\\
\end{array} \\
\quad & \,\,\, g_{10} = \text{min}(\Delta x_{\text{si, i = 1, 2, ..., n-1 }}) \geq \text{1e-03}, & \\
\quad & \text{Velocity constraints: } \\
& \begin{array}{ll}
g_{11} = \dot{x}_{\text{s1\_E}} > 0, & g_{12} = \dot{y}_{\text{s1\_E}} > 0,\\
g_{13} = \dot{y}_{\text{s1\_E}} < 5, & g_{14} = S_{\text{s1}} < 0.3\\
g_{15} = \dot{x}_{\text{s2\_E}} > 0, & g_{16} = \dot{y}_{\text{s2\_E}} > 0,\\
g_{17} = \dot{y}_{\text{s2\_E}} < 5, & g_{18} = S_{\text{s2}} < 0.3\\
\end{array} \\
\quad & \text{Rotation constraints: } \\
& \begin{array}{ll}
 g_{19} = \delta\omega_{\text{f1}} > \pi, & g_{20} = \delta\omega_{\text{f1}} < 2\pi \\
\end{array} \\
\quad & \text{Additional constraints: } \\
& \begin{array}{lll}
g_{21} = T_{\text{1}} > \frac{1}{15}, & g_{22} = T_{\text{1}} < 2 , &
g_{23} = N \geq 8 \\
\end{array} \\
{\text{where}:}\quad & \mathbf{X} = [\text{Motor Label}, m_{add}, E, \rho, \text{b}, h, b_l, \dot{\zeta}_0, \\& \theta_0, \dot{\theta}_0, a_{i, i=0,1, ..., 5}, t_{\text{FC}}]
\end{aligned}
\end{equation}

In the equation, \(\theta_{\text{Diff}}\) is the absolute value of the designed variable initial touchdown angle, \(\theta_0\), and the second cycle's touchdown angle, \(\theta_{\text{TD2}}\). The upper and lower bounds of the input variables are defined by \(\mathbf{X}_{\text{L}}\) and \(\mathbf{X}_{\text{L}}\) respectively. All constraints were weighted equally, and the definitions of the notations in the constraints shown in Eq. \eqref{eq:obj_mdpso_angle_diff} are explained below:
\begin{itemize}
\item \textbf{Touchdown angle constraints: } \(\epsilon\) is the standard deviation of the Gaussian noise adding to the initial touchdown angle. 

\(g_1\) and \(g_3\) bounds the touchdown angles of the first two stance phases within the pre-defined initial condition. \(g_2\) ensures the difference between the two touchdown angles is smaller than the standard deviation of the Gaussian noise, which instructs the robot to return to the same range of the touchdown angle as at the beginning of the simulation.
\item \textbf{Position constraints:} $y_{\text{s1\_S}}$, \(y_{\text{s1\_M}}\) and \(y_{\text{s1\_E}}\) represents the \(y\) coordinate of the first stance phase's start point, midpoint, and endpoint respectively. \(\mathbf{x}_{\text{s1}}\), \(\mathbf{y}_{\text{s1}}\) are arrays of the $x$ and $y$ coordinates of the first stance phase, and \(\mathbf{y}_{\text{s2}}\) is the array of the $y$ coordinates of the second stance phase. \(\Delta x_{\text{f1}}\) is the $x$ direction displacement during the first flight phase. \(\Delta x_{\text{si, i = 1, 2, ..., n-1 }}\) is the $x$ direction displacement of each stance phase from the first phase to the second to last phase, where n is the total number of phases completed in the simulation.

\(g_4\) and \(g_5\) ensure the leg compresses and decompresses during stance, resulting in SLIP dynamics, as opposed to staying rigid and resulting in a vaulting motion. \(g_6\) prevents the robot from ending in a position backward of its starting position. \(g_7\) and \(g_9\) ensure the position of the robot's center of mass does not fall below the terrain for the first two stance phases. \(g_8\) constrains the distance traveled during flight to mimic travel distances seen in biological runners~\cite{Stuhec:2023aa}. \(g_{10}\) was applied to ensure a minimal travel distance during stance, which prevented the system from converging towards a hopping gait without forward movement.

\item \textbf{Velocity constraints:} $g_{11}$ and $g_{15}$ ensure that the direction of travel is in the positive $x$ direction at the end of the first two stance phases.  $g_{12}$, $g_{13}$, $g_{16}$, and $g_{17}$ specify the minimum and maximum bounds on the $y$ velocity at the end of the stance phase.  This velocity must be nonzero to ensure a flight phase, and the upper bound prevents excessive and unrealistic jump heights. 

Legged locomotion is a periodic dynamical system. To achieve a stable gait, the initial conditions of the first step must match those of the subsequent steps. This results in a symmetric position and velocity vector during stance. Constraints $g_{14}$ and $g_{18}$ apply upper bounds on the symmetry of the velocity vectors for the first two stance phases where $S$ is defined in \eqref{eq:sym}.  Velocity was chosen for the constraint as positional or trajectory symmetry does not guarantee the symmetry in velocity magnitude. However, a symmetry in velocity magnitude can imply a symmetry in position or trajectory in SLIP.
\begin{equation}
\label{eq:sym}
S=\left(1-\frac{\dot{x}_{\text{si\_E}}}{\dot{x}_{\text{si\_S}}}\right)^2+\left(1-\frac{\dot{y}_{\text{si\_E}}}{\dot{y}_{\text{si\_S}}}\right)^2
\end{equation}
Equation \eqref{eq:sym} specifies the change in normalized $x$ and $y$ velocities from the start to the end of the stance.  The smaller the value of $S$, the more symmetric the gait is.



\item \textbf{Rotation constraints:} \(g_{19}\) and \(g_{20}\) limit the robot's rotation in the flight phase within \(180^\circ\) to \(360^\circ\). This ensures the leg rotates during the flight phase to reposition for the following phase but does not over rotate.

\item \textbf{Additional constraints:} \(T_1\) is the period for the first cycle. \(g_{21}\) and \(g_{22}\) bound $T_1$ in the range  $[1/15,2]$ seconds, which matches stride periods seen in biological runners, and would be seen in robots with similar dynamics~\cite{Full1990,Heglund1988SpeedSF}. 

The minimum number of cycles $N$ was constrained by \(g_{23}\) to be greater than 8.  As constraints were only applied to the first two cycles, this ensures the system is able to complete additional steps.
\end{itemize}

\subsection{Case Study 2: Optimizing Energy Consumption}
Minimizing energy usage due to actuation can be calculated by integrating the power consumed by the motor over one stance and flight cycle which is described by equation \eqref{cost_func_main}.
\begin{equation}
\label{cost_func_main}
F=\int_{0}^{t} V_a(t) \cdot i_a(t) \,dt
\end{equation}
The objective function in this problem is defined as equation \ref{eq:obj_MDPSO_energy}
\begin{equation}
\label{eq:obj_MDPSO_energy}
\min_{\mathbf{X}} \quad  f(\mathbf{X}) = F(\mathbf{X})
\end{equation}
Constraints between the two case studies are the same except for \(g_3\), \(g_8\), and \(g_{24}\), which are shown in the following equations:
\begin{equation}
\label{eq:c2_cons}
\begin{aligned}
g_2 =  |\theta_0 - \theta_{\text{TD2}}(\mathbf{X})| < 0.859\\
g_{8} = 5 \cdot (|\Delta x_{\text{f1}}| - 4 \cdot l_0) < 0\\
g_{24} = -0.001 - \min (\mathbf{V}_{\text{s1}} \cdot \mathbf{I}_{\text{s1}}) < 0 
\end{aligned}
\end{equation}
The new \(g_3\) constraint is defined based on the result of optimizing the touchdown angle difference, which will be discussed in the next section. The value of \(0.859^{\circ}\)  is approximately two times the converged touchdown angle difference from case study 1.  Therefore, this second case study is encouraged to achieve similar stable results as the previous optimization. The new \(g_8\) was weighted by a factor of 5 as it was the most difficult constraint to satisfy. This weighting factor helps MDPSO prioritize the constraint. \(g_{24}\) was added to prevent the system from stalling but was not needed when optimizing the touchdown angle. A slightly negative value was used to help with numerical instability.
\subsection{Optimization Parameter Bounds}

\begin{table}
\caption{Optimization Variables}
\label{tab:opt_param}
\begin{center}
\begin{tabular}{l| c | c | c c }
Parameter & units& Type & Min & Max\\
\hline
Motor Label    & -             &Int   & 1      & 18\\
$m_{\text{add}}$      & \unit{kg}     &Float & 0.01      & 0.5\\
E              & \unit{Pa}     &Int & 10e3   & 130e9\\
$\rho$         & \unit{m}      &Float & 0.001  & 0.03\\
b              & \unit{m}      &Float & 0.0005  & 0.01\\
h              & \unit{m}      &Float & 0.0005  & 0.01\\
$b_l$          & \unit{Ns/m}   &Float & 0.0005   & 0.01\\
$\dot{\zeta}_0$  & \unit{m/s}  &Float & -5    & -0.1\\
$\theta_0$       & \unit{deg}  &Float & 25    & 85\\
$\dot{\theta}_0$ & \unit{RPS}&Float & 0.8      & 1.6\\
\edit{$T_{\text{FC}}$}  & \unit{s}    &Float & 0 & 0.1\\
$a_5-a_0$        & -           &Float & -2 & 2\\

\end{tabular}
\end{center}
\vspace{-15pt}
\end{table}

The optimization framework for component choice and design requires setting feasible physical bounds. These bounds are detailed in this section.
\begin{itemize}
    \item \textbf{Motor Label} Equations \eqref{motor_cur} and \eqref{DC_motor_torq} show the parameters that affect motor performance, which cannot be independently varied without designing new motors. \edit{As such, the motor input is a discretized list where each value corresponds to a different motor and gearbox combination. For this work the \SI{3}{\volt} brushed \SI{6}{\milli\meter}, \SI{8}{\milli\meter}, and \SI{10}{\milli\meter} motors from Maxon (maxongroup.com) were considered.  The \SI{3}{\volt} option was chosen as they can be powered from a single-cell lithium polymer battery, which typically operates at \SI{3.7}{\volt}.  When the three motor sizes are paired with their corresponding gearbox options, the optimizer is presented with 18 options to choose from.}


    \item $\mathbf{m_{add}}$ is the additional mass added to the system on top of the minimum required mass.  Here the minimum mass is estimated as two motors ($m_{motor}$), a microcontroller development board estimated at \SI{5}{\gram} ($m_{mcu}$), and a battery mass ($m_B$) of \SI{3}{\gram} which represents a \SI{3.7}{\volt}, \SI{100}{\milli\ampere\hour} lithium polymer battery. This minimum mass is multiplied by two to estimate the supporting structure mass, $m_{min} = 2 (m_B + m_{mcu} + 2 m_{motor})$.

    
    \item \textbf{E} The bounds of the elastic modulus correspond to castable polymers (\SI{10}{\kilo\pascal}) to brass (\SI{130}{\giga\pascal}). A material will have to be chosen based on the optimized value since a discrete option is not simulated. 
    \item $\bm{\rho}$,\textbf{b,h} correspond to the leg properties. A lower bound of \SI{0.5}{\milli\meter} was chosen since this corresponds to 26 gauge sheet metal or a few layers on most 3D printers. In hardware, geometry will need to be refined to accommodate both modulus mismatch and manufacturing limitations. 
    
    \item $\mathbf{b_l}$ It is difficult to design specific damping without extensive testing of leg materials and geometries. Damping is included in the model since damping improved stability in legged locomotion~\cite{damping_important,stpierre2020viscoelastic}, though damping would have to be experimentally characterized. 
    

    \item ${\dot{\mathbf{\zeta}_0},\mathbf{\theta_0},\dot{\mathbf{\theta}_0}}$ correspond to the systems initial conditions. $\dot{\zeta}_0$ must be negative for the leg to compress. Bounds on $\theta_0$ were chosen based on observed values in biology and robotics, which tend to fall within the range of $47^{\circ}$ to $82^{\circ}$~\cite{Muller:2016,non_linear_k,SEYFARTH2002649,SLIP_learning,self_stable,swing-leg}. To get more flexibility in the search space, the lower bound on \(\theta_0\) is set to $25^{\circ}$. Similarly, bounds on $\dot{\theta}$ were chosen based on biological data and speeds of similarly sized robots~\cite{saranli2001rhex,barragan2018minirhex,c-quad,stpierre2020viscoelastic}. For example, cockroaches (\SI{2.6}{\gram}) and horses (\SI{680}{\kilo\gram}) have gait frequencies \SI{15}{Hz} and and \SI{2}{Hz}, respectively~\cite{Full1990,Heglund1988SpeedSF}. As the leg in the proposed design will complete one rotation per stride, this correlates to a rotational speed of 2-15 RPS.  These bounds on $\theta$ and $\dot{\theta}$ impact motor choice as they affect the torques and speeds the motor must supply to the system.

    \item $\mathbf{T_{FC}}$ \edit{represents the time period the control voltage will be applied during the stance phase.  Using the bounds placed on $\dot{\theta_0}$ above, the system will operate in the range of 2-15~\SI{}{Hz}, which correlates to a flight cycle time of less than \SI{0.5}{s}.}

    \item $\mathbf{a_5 - a_0}$ The bounds on these polynomial coefficients were chosen heuristically to be as generous as possible without imposing an unnecessary constraint. These bounds do not limit the maximum voltage, which is handled by constraint \(g_2\).   
    
\end{itemize}

\begin{table}
\caption{Optimized Solutions}
\label{tab:opt_sol}
\begin{center}
\begin{tabular}{c| c| c  }
Variables & Case Study 1 & Case Study 2 \\
          & Minimizing \(\theta_{\text{Diff}}\) & Minimizing \(F\)\\
\hline
Motor Label      & 15             &15       \\
$m_{\text{add}}$ & 0.400          &0.330 \\
E                & 1.173e+09      &3.133e+08 \\
$\rho$           & 0.0300         & 0.0176   \\
b                & 0.0100         & 0.00301 \\
h                & 0.00425        & 0.00717 \\
$b_l$            & 0.00916        & 0.00683 \\
$\dot{\zeta}_0$  & -1.0467        & -1.098   \\
$\theta_0$       & 69.385          & 63.942    \\
$\dot{\theta}_0$ & 1.796        & 4.104   \\
$a_0$            & 1.383          & \editt{1.912}   \\
$a_1$            & 1.0727         & \editt{-0.0466}   \\
$a_2$            & 2.000          & \editt{-1.154}   \\
$a_3$            & -1.441         & \editt{-1.936}   \\
$a_4$            & 1.706          & \editt{-0.167}   \\
$a_5$            & -0.398         & \editt{1.102}   \\
$T_{\text{FC}}$  & 0.0740         & 0.0701   \\
\hdashline
$\theta_{\text{Diff}}$ & 1.053   & 1.053  \\
$F$                    & 0.125   & 0.0777 \\
\end{tabular}
\end{center}
\end{table}

\section{Optimized Designs: Results \& Discussion \label{sec:results}}
The MDPSO optimization process was performed on an AMD Ryzen 9 5950X 16-Core Processor CPU with 64 GB RAM Windows workstation. The population size of MDPSO was set to 128, and the optimization ran with 32 parallel workers in MATLAB. Optimized design and control parameters are listed in Table \ref{tab:opt_sol} for both case studies, and Figure \ref{fig:renderings} shows a rendering of the optimum designs resulting from the two cases. The following termination criteria were used in the optimizations: \textbf{1)} the minimum objective is infeasible and the net constraint violation does not decrease for 15 consecutive iterations; \textbf{2)} the minimum feasible objective does not decrease for 5 consecutive iterations.  The case of minimizing touchdown angle difference finished in 0.93 hours with 47 iterations, and the optimization of minimizing energy cost finished in 1.58 hours with 95 iterations. The convergence histories of the two case studies are shown in Figure \ref{fig:conv_td} and Figure \ref{fig:conv_e}. Case Study 2 took longer to converge most likely since finding feasible solutions took relatively greater number of iterations, as seen from Figure \ref{fig:conv_e}. The convergence history plots show that the objectives of the two case studies converged to \(0.446^{\circ}\) and \SI{9.81}{mJ} respectively, but due to the Gaussian noise added to the first touchdown angle, the converged objectives still need to be validated.

\begin{figure}[t]
   \centering
   \includegraphics[width=0.7\linewidth]{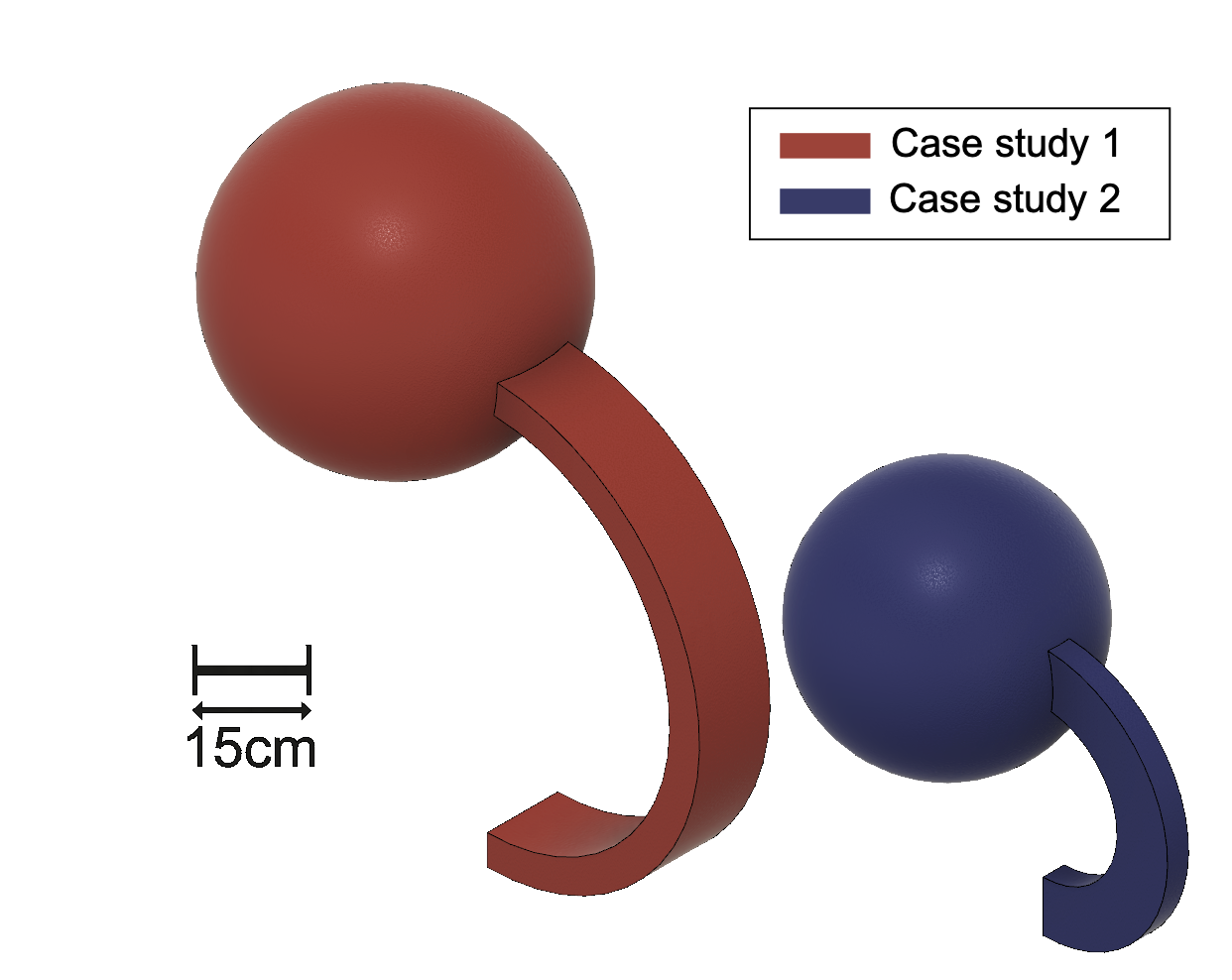}
   \caption{Renderings of optimized designs. Red: case study 1 -- minimizing \(\theta_{\text{Diff}}\); and Blue: case study 2 -- minimizing \(F\). The sphere diameter is indicative of the relative system mass.}
   \label{fig:renderings}
   \vspace{-10pt}
\end{figure}
\begin{figure}[t]
   \centering
   \includegraphics[width=0.9\linewidth]{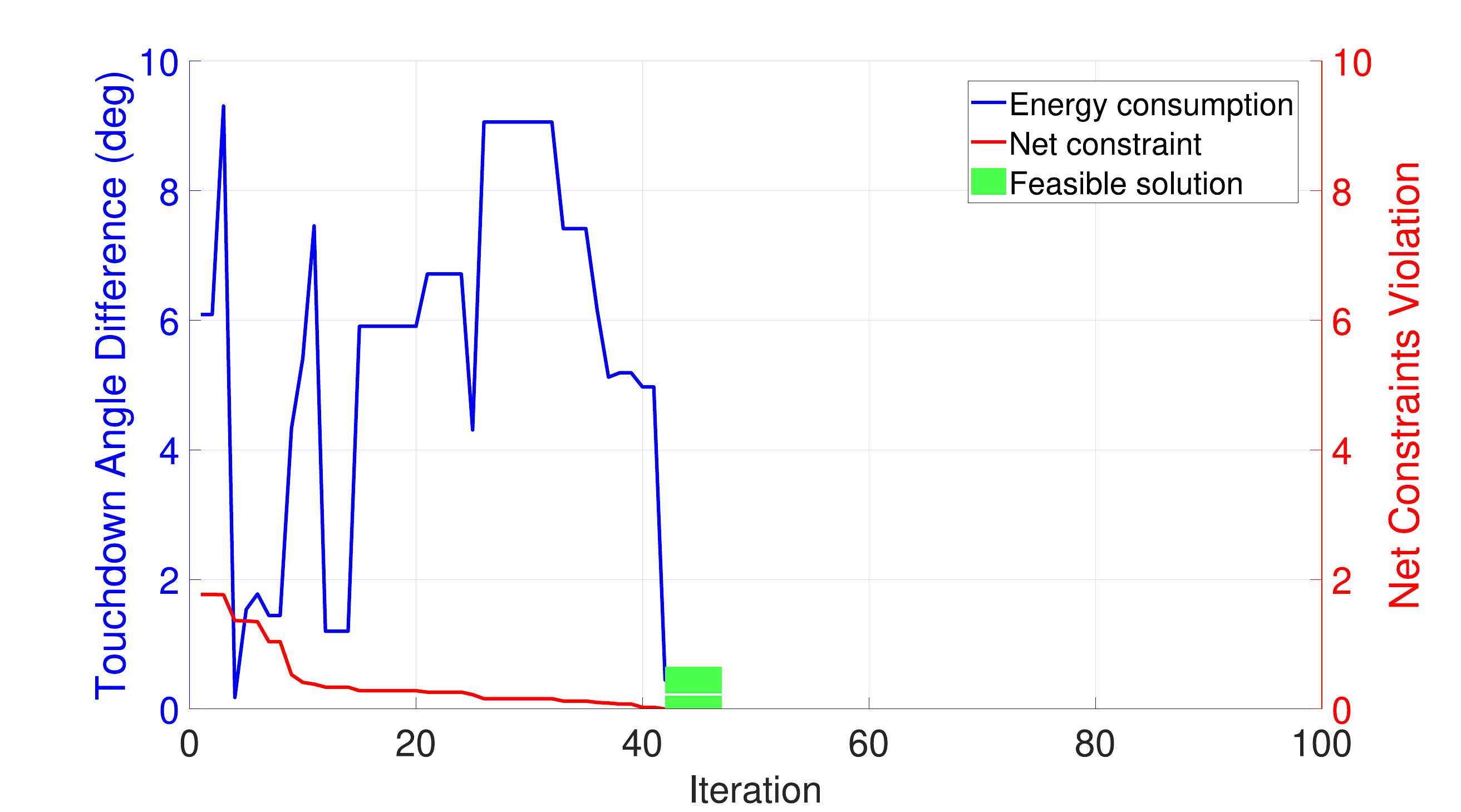}
   \caption{Cost function and constraint convergence history for case study 1 where variation in touchdown angle was minimized.  Feasible results are highlighted in green.}
   \label{fig:conv_td}
   \vspace{-10pt}
\end{figure}
\begin{figure}[t!]
   \centering
   \includegraphics[width=0.9\linewidth]{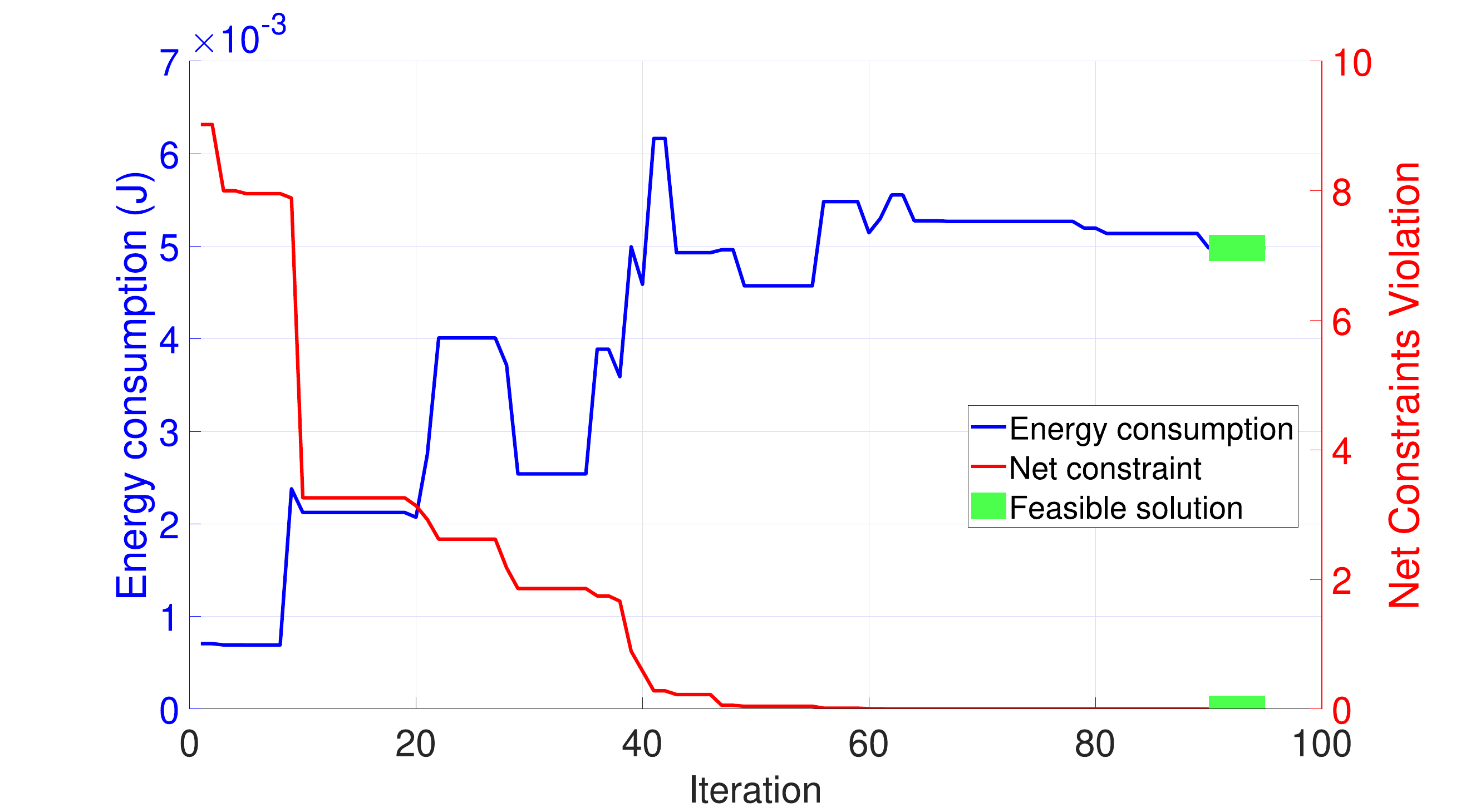}
   \caption{Cost function and constraint convergence history for case study 2 where input energy was minimized.  Feasible results are highlighted in green.}
   \label{fig:conv_e}
   \vspace{-10pt}
\end{figure}

To validate the converged objectives of the two optimizations and to quantify the robustness to uncertainty, designs from both case studies were simulated 100 times with noise on the optimized touchdown angle. In these evaluations, the noise was uniquely instantiated for each sampling, following a normal distribution denoted by \(\mathcal{N}(0, \epsilon^2)\).  Both case studies were validated with the same set of noise, and the touchdown angle difference, \(\theta_{\text{Diff}}\), was \(1.053^{\circ}\), which is within \(\epsilon\) of \(1.2896^{\circ}\) that defines the Gaussian noise during optimization. The result shows that case study 2 can achieve the same \(\theta_{\text{Diff}}\), due to the constraint \(g_2\), which was defined based on the converged \(\theta_{\text{Diff}}\) of the first case study. The value 1.053 is slightly greater than the threshold set in the constraint due to the noise, but it shows using the result of the first case study to define the constraint can encourage the optimizer to achieve stable gaits. This addition of noise in both case studies verifies that the optimization approach here could result in robots that are robust to noise in hardware realization or during operation. The energy cost of case study 2 is much smaller than that of case study 1, which shows that given the same motor selection, the energy cost can be reduced with tuning of the physical design parameters and control voltage profile.

In addition to evaluating the standard deviation of the touchdown angle difference, the number of gait cycles was quantified during this validation step. In case study one, the average cycle completed before the system was no longer stable was 5, and the maximum number of cycles completed in a single evaluation was 12 cycles, failing during the last flight phase. In Case Study 2, the average cycles completed in each evaluation was 6.5, which is a step and a half greater than the first case study, and the maximum cycles completed in a single evaluation was 20. Despite the higher variance in touchdown angle difference, the optimized design, which minimized energy, results in more stable dynamics, completing more gait cycles in comparison to the optimized design in case study one.

Figure~\ref{fig:COM} plots the trajectories of the optimized designs during this validation step. The trajectories show similar dynamics, largely as a result of the similar system parameters found in each optimization case study. For example, both used motor option 15, which corresponds to Maxon's \SI{10}{mm} diameter motor PN 118383 with a gear reduction of 16. While the overall system sizes, i.e., masses and lengths, are different, resulting in calculated leg stiffnesses of \SI{1769}{\newton\per\meter} and \SI{3382}{\newton\per\meter}, receptively, their relative stiffnesses are similar. The corresponding relative stiffnesses ($k_{rel} = \frac{k_0l_0}{mg}$) were found to be 12 and 16, implying similar dynamics between the two systems, and falling within the stable regimes reported in~\cite{SHEN2015433}.


The open-loop control voltages are shown in Figure~\ref{fig:Control}, corresponding to the trajectories shown in Figure~\ref{fig:COM}.  As the control is an open-loop profile repeating for each stance and flight cycle, the control was only plotted through the second stance phase.  While the control profile was specified as a fifth-degree polynomial, the optimization returned a piece-wise linear profile, which is easier to implement in hardware. This profile is similar to the Buehler clock utilized in the RHex robot~\cite{bclock}, but with an added delay during flight. It is possible that this strategy is a robust open-loop strategy that can be used in a diversity of robots.

\begin{figure}[h]
    \centering
    \includegraphics[width=0.9\linewidth]{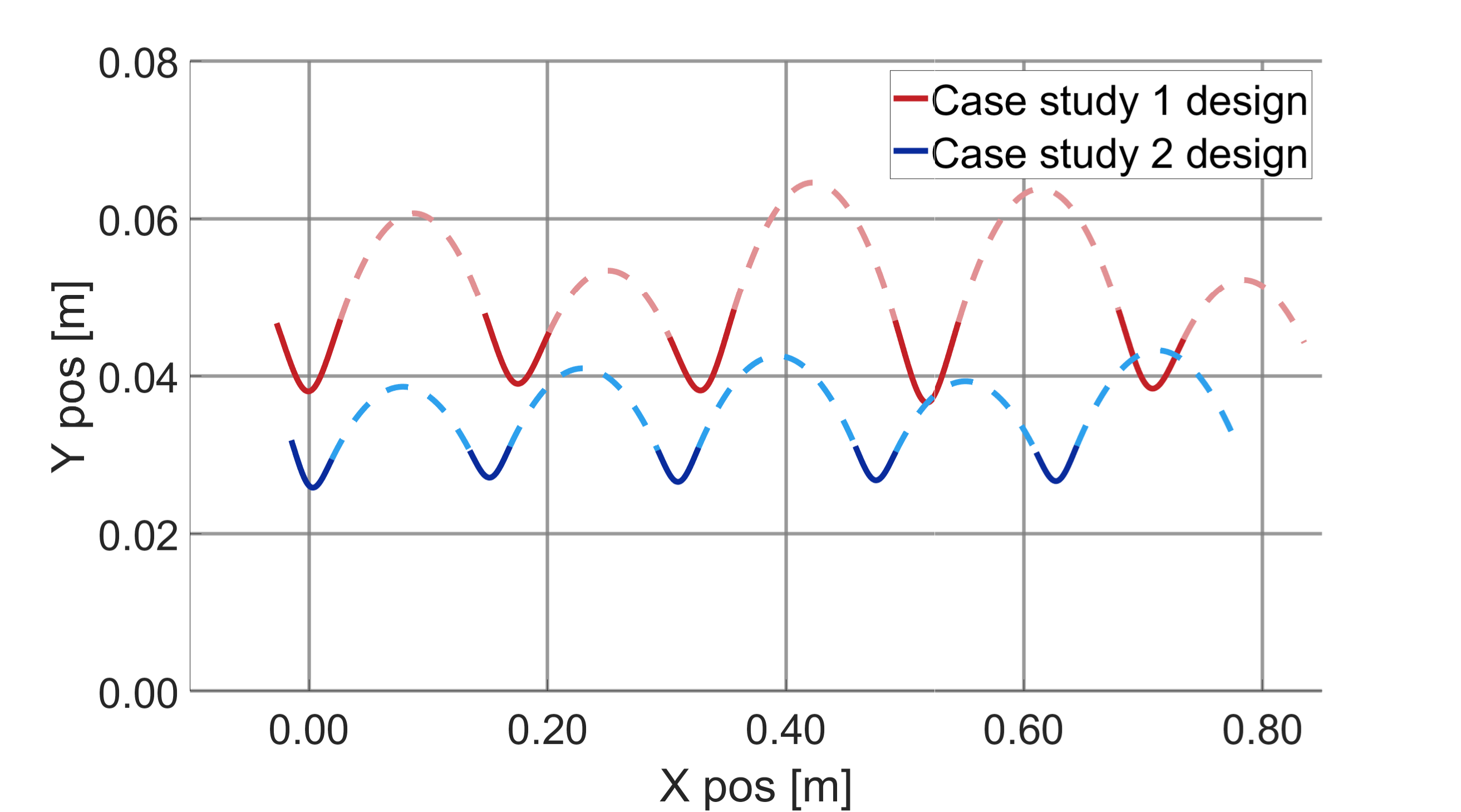}
    \caption{Simulated trajectories of select optimized solutions. Solid lines denote stance phases and dashed denote flight phases.}
    \label{fig:COM}
\end{figure}
\begin{figure}[h]
    \centering
    \includegraphics[width=0.9\linewidth]{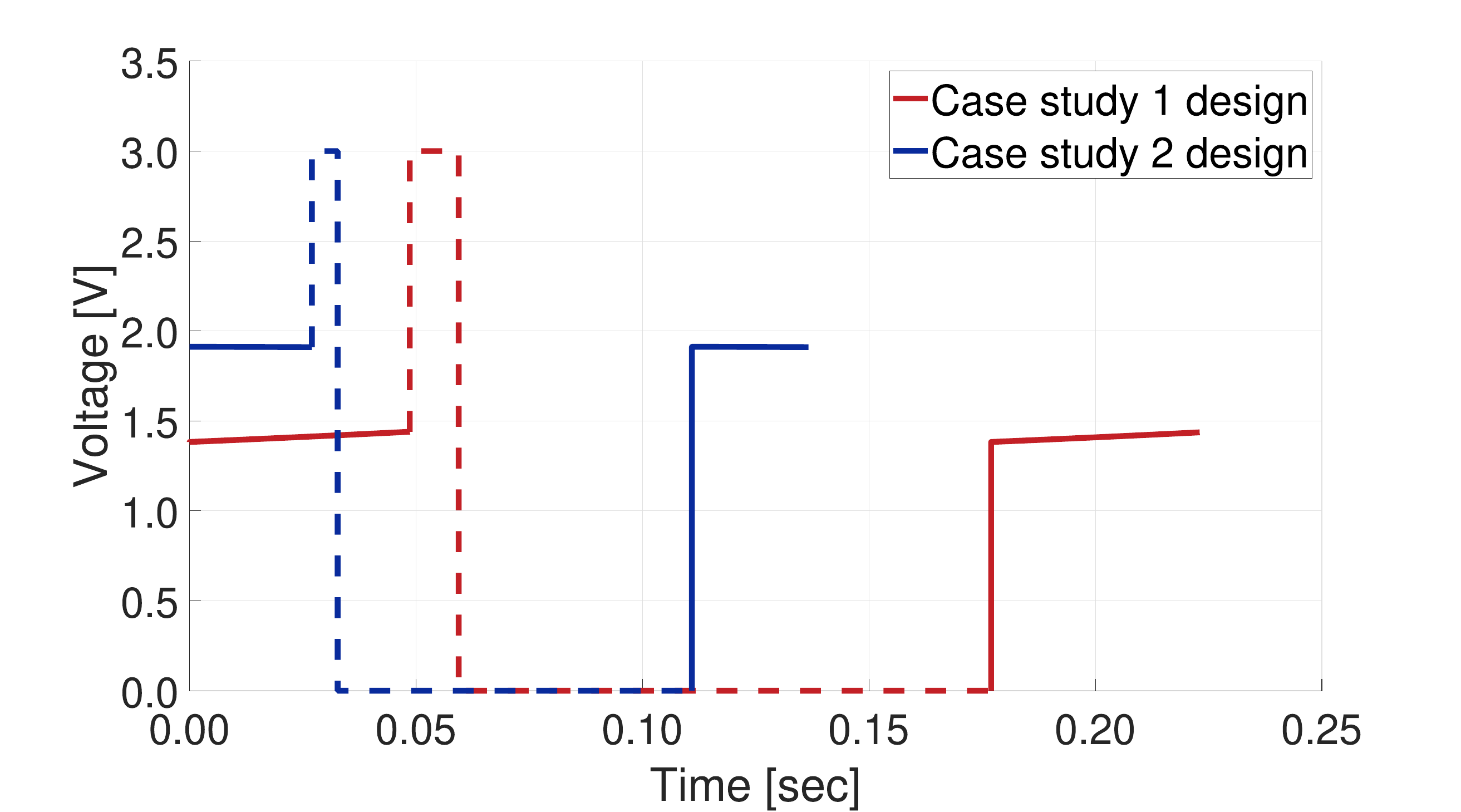}
    \caption{Control voltage for the trajectories shown in Figure~\ref{fig:COM}. Solid lines denote stance phases and dashed denote flight phases.}
    \label{fig:Control}
    \vspace{-15pt}
\end{figure}



\section{Conclusions \label{sec:conclusions}}
In this paper, we present an efficient computational framework to explore optimized designs of legged robots with the torque-driven spring-loaded inverted pendulum (TD-SLIP) abstraction operating on flat terrain. Two case studies were performed, the respective objectives of which involved optimizing the touchdown angle difference and the energy consumption between the first two cycles. Hardware and physical parameters such as motor selection, leg geometry, and representative mass were concurrently considered alongside control parameters, such as voltage profiles and touchdown angles, in the optimized design process. 
%
The optimized designs obtained in this work were observed to adhere to the current understandings of the dynamics of legged locomotion (albeit subject to the assumed fidelity of the modeling process); the optimized designs presented relative stiffnesses in the range of biological and robotic runners~\cite{SHEN2015433}, providing initial evidence for the suitability of the set of constraints formulated to drive the design process while preserving a degree of realism. These designs were observed to be robust against a $3^{\circ}$ noise in the first touchdown angle with a 98\% confidence interval. When tested over 100 validation runs, these open loop designs completed over five (flight/stance) cycles on average. 

There remains scope to further improve the repeatability of the TD-SLIP motion; potential future extensions in this regard include closed loop controls, and efficient uncertainty propagation techniques to impose reliability constraints. In addition, higher fidelity analysis or physical testing of the leg's structural dynamics and terrain interaction in the future could provide further insights into both the effectiveness of the proposed design framework and the achievable capabilities for such small-scale running robots. 






\section*{Acknowledgments}
This work was supported by the startup funds provided by the Department of Mechanical and Aerospace Engineering and the School of Engineering and Applied Sciences at the University at Buffalo, and the National Science Foundation (NSF) award CMMI 2048020.

\bibliographystyle{IEEEtran}
\bibliography{sample} 


\end{document}